# Intelligent Motion Planning for a Cost-effective Object Follower Mobile Robotic System with Obstacle Avoidance

Sai Nikhil Gona[1], Prithvi Raj Bandhakavi[2]

Electrical and Electronics Engineering[1], Mechanical Engineering[2]

Chaitanya Bharathi Institute of Technology, Telangana, India[1,2]

E- mail: ugs16047_eee.sai@cbit.org.in [1], pbandha@ncsu.edu [2]

*Abstract*— There are few industries which use manually controlled robots for carrying material and this cannot be used all the time in all the places. So, it is very tranquil to have robots which can follow a specific human by following the unique coloured object held by that person. So, we propose a robotic system which uses robot vision and deep learning to get the required linear and angular velocities which are v and ω, respectively. Which in turn makes the robot to avoid obstacles when following the unique coloured object held by the human. The novel methodology that we are proposing is accurate in detecting the position of the unique coloured object in any kind of lighting and tells us the horizontal pixel value where the robot is present and also tells if the object is close to or far from the robot. Moreover, the artificial neural networks that we have used in this problem gave us a meagre error in linear and angular velocity prediction and the PI controller which was used to control the linear and angular velocities, which in turn controls the position of the robot gave us impressive results and this methodology outperforms all other methodologies.

*Key words* – Computer Vision; Artificial Neural Networks for Computing Linear and Angular velocity; PI Controller for Velocity Control; Object Follower Robotic System with Obstacle Avoidance; Colour Spaces based Image Segmentation.

*Nomenclature* – ANNs – Artificial Neural Networks, RGB-Red Green Blue, HSV- Hue Saturation Value, MSE- Mean Squared Error, MAE – Mean Absolute Error

1. INTRODUCTION

The field of artificial intelligence and robotics is playing an imperative role in our lives these days [1], human efforts have abridged a lot due to artificial intelligence and researches are still being done in artificial intelligence [2] to make household and domestic works simpler with minimal physical efforts. Artificial neural networks with computer vision are one of the most important techniques to produce intelligent domestic and industrial robots. Carrying weights from one place to the other is where a lot of human efforts are required and the labour cost has increased a lot in the past couple of years so, it is getting very difficult for the industrialists to hire more labour in industries, airports and shopping malls for carrying weights from one place to the other. So, in order to reduce the cost spent on labour, we need to adopt an automatic mechanism instead of employing a lot of people as labours. The most important factor what we need to consider while adopting an automatic mechanism is the cost of the mechanism. In this kind of scenario, we can just adopt mobile robotic systems which can do the work efficiently, which is to carry heavy objects from one place to the other and this brings down the amount of money spent on the labour and the overall money spent on manufacturing.

In the US, the shortage of labour has been a substantial problem in the industries and 8 out of 10 manufacturing executives feel that workforce shortages or skill deficiencies in production roles have a negative and a significant impact on their ability to meet the customer demand. For this reason, industries have adopted artificial intelligence techniques, intelligent systems, and robotics. Any work in the industrial sector which requires a lot of labour can be replaced by robots. Adopting robots also reduce the amount of money spent on the workforce since robots are developed in a such a way that the work done by one robot is equivalent to the work done by more than one individual. So, to meet customer demands, the role of robotics and artificial intelligence will be significant. For this reason, we propose to develop a solution using an intelligent mobile robotic system for material carrying problems in the industries.

In this paper, we propose a cost-effective mobile robotic system which carries loads [3] and follows a specific person by detecting the unique coloured object held by that person, by avoiding obstacles using artificial neural networks and computer vision [4]. We have made an algorithm, inspired by the colour-based object identification algorithm, for detecting the position of the unique coloured object. We get the position of the unique coloured object with the help of camera [4] and the position of the obstacles with the help of the ultrasonic sensors placed at 2 different positions in the robot. We used Beagle Bone Blackboard interfaced with camera and sensors, and we have used "TensorFlow" which is an open-source software or library to build the artificial neural networks to get the desired linear velocity and desired angular velocity required by the robot to move its left and right wheel. We have tested our methodology on a quick bot with 2 driven wheels and one free driving wheel. We can apply the same methodology on humanoid robots to follow a human by avoiding obstacles. This could replace the normal shopping carts in the malls since it needs no force to be applied to it for motion and the whole process is automated. The most common controller that everyone uses is a PID controller, the abbreviation of PID controller is proportional, integral, and derivative controller, for this particular problem we are using a PI controller since we are only controlling the robot's linear and angular velocities.

The major contributions of this paper are as follows:

(1). A new methodology inspired by the standard colour-based object detection method, which tracks the position of the object horizontally and computes if the object is close to the robot or far from the robot.

(2). A hybrid cost-effective approach of combining the real-time colour-based object detection and sensor readings with the standard artificial neural networks to compute linear and angular velocities while following the object and avoiding the obstacles. This eliminates the need of CNNs and a GPU for person recognition and tracking which makes the model cost-effective.

Section 2 provides an overview of the existing object detection and follower mobile robotic systems. The proposed methodology is given in section 3 in detail. Section 4 provides an overview of the hardware of the robot and the functions used to calculate the position of the robot from the ticks of the wheel encoders. Section 5 provides a precise analysis of the results obtained from the proposed methodology. A brief outline of the current work and its future scope is given in section 6.

## 2. RELATED WORKS

Several works on object tracking can be found in the present literature, most of the existing works are only focused on object tracking without obstacle avoidance. In [5] a methodology was proposed in which the standard colour based object detection algorithm has been used, in this method, the robot identifies the midpoints of the object and follows the largest object that it has tracked and it doesn't avoid obstacles. The robot might miss the actual object, that needs to be followed, in the course of following the largest object. In [6] a methodology was proposed which uses the standard colour based object tracking algorithm, in this method the robot identifies the object and the robot moves only depending upon the position of the ball in the video frame irrespective of the distance between the ball and the robot, this method is unreliable since the motion of the robot very much depends on the position of the object in the axis perpendicular to the robot which tells us if the object is close to or far from the robot and this robot doesn't avoid obstacles. In [7] the proposed robot design tracks the object based on colour using adaptive colour matching and Kalman filter. It also tracks obstacles by forming an obstacle cluster, but this method does not hold good while dealing with transparent obstacles. In [8] the proposed methodology tracks the object using Lucas and Kanade's algorithm and the obstacles are avoided by making a new path every time the goal configuration is changed, using dynamic goal potential fields algorithm. In [9] a method was proposed in which the object is detected after converting the video to a binary one, then the threshold value will be selected in the range between 2 and 211 of grayscale and then the objected is isolated and tracked, this method doesn't avoid obstacles. In [10] a multicoloured tag is detected based on the orientation of the colours in the tag and the ultrasonic sensors give the distance between the tag and the robot, if the distance is more than a threshold value the robot moves forward, this robot cannot avoid obstacles. The novel method that we are going to propose will track the position of the ball in the horizontal axis and it also tells us if the object is close to or far away from the robot, unlike [6], which doesn't care if the ball is close or far. Unlike [5] this method doesn't follow the largest object, it only follows the object that needs to be followed. The proposed method plans the mobility of the robot intelligently by avoiding obstacles using ANNs, unlike [5][6][8][9][10] in which the robots don't avoid the obstacles and since this method is planning the mobility of the robot, the capability of the robot to adjust itself towards the object during the sudden appearance of the obstacles is quicker than the method proposed in[8]. Unlike the method proposed in [7], this method also avoids transparent obstacles as we have used ultrasonic sensors for detecting the obstacles. The vertical position of the object is not required since the robot only needs to go either left or right or straight depending upon the horizontal position of the ball. This method avoids obstacles quickly, intelligently and plans the movements of the robot and this model has no drawbacks that were there with the previous models.

## 3. PROPOSED METHODOLOGY

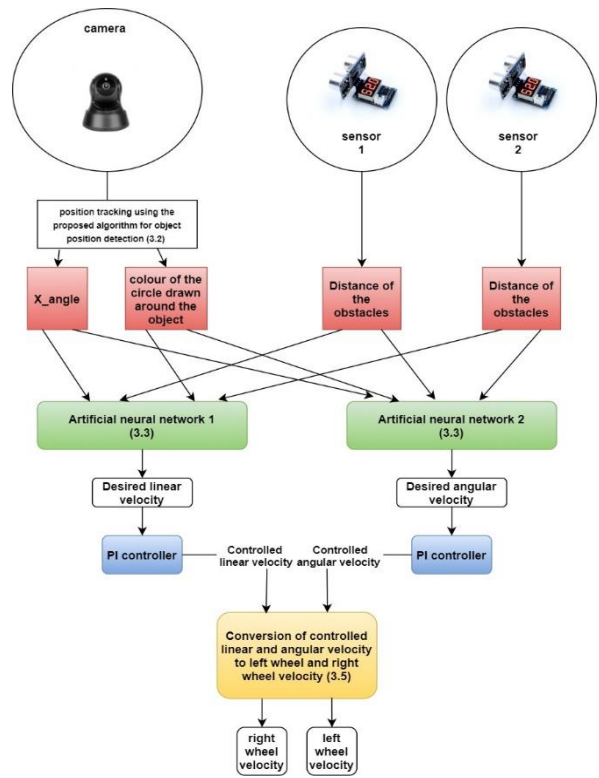

**Fig. 1. Proposed methodology for mobility planning**

3.1. Explanation

Fig. 1 shows us the complete architecture of the proposed methodology, the methodology that we proposed has 3 stages. The first stage is that stage where we find the $x_{angle}$, the distance of the obstacles from the robot using sensor 1 and 2, and the closeness of the object to the robot, by the colour of the circle drawn around the object. Now comes the second stage where we take the values of the $x_{angle}$, the colour of the circle drawn and the values from both the sensors and feed them to the artificial neural networks. Sensor 1 and sensor 2 can also be called as 'left sensor' and 'right sensor', respectively, as sensor 1 was placed to the left side of the robot and sensor 2 was placed to the right side of the robot. We have used artificial neural networks for computing linear velocity and angular velocity with which the robot should move and in the third stage, the linear and angular velocities are given to the PI controller which gives us the controlled linear and angular velocities, the controlled linear velocity and angular velocity are then converted into the velocity of the right wheel and the left wheel for the robot to move. The proposed methodology generates the most optimal linear velocity and angular velocity in every time step and the process of generation of linear and angular velocities is quick enough to react to the sudden appearance of obstacles, which makes the methodology novel and better than the previous methodologies of object follower with obstacle avoidance mentioned in section 2.

We do not require the $y_{angle}$ since the robot needs to move either left or right or straight and $y_{angle}$ is only required when the robot needs to move up or down according to the altitude of the object that needs to be followed. In 3.2, 3.3, 3.4 and 3.5 we are going to discuss the proposed methodology as follows:

(1). In 3.2 we discuss the novel algorithm that we have used for detecting the object and tracking its position. This algorithm takes in the video as an input and produces the object's position horizontally and the colour of the circle around the object.

(2). In 3.3 we discuss the artificial neural networks used in the proposed novel methodology to predict linear and angular velocities. The sensor readings, horizontal position of the object in pixels and the colour of the circle around the object are fed into the neural networks and the linear and angular velocities are given out as outputs.

(3). In 3.4 we describe the design of the PI controller for controlling the linear and angular velocities. The PI controller computes the controlled linear and angular velocities, which controls the robot depending upon its present state.

4). In 3.5 we show how the linear and angular velocities are converted into left wheel velocity and right velocity.

3.2. The algorithm used for detecting the object's position-

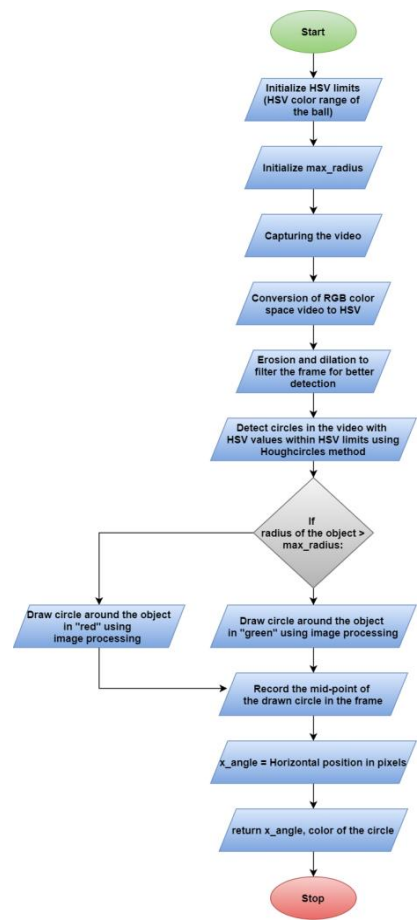

**Fig. 2. The algorithm used for object position detection**

Fig. 2 shows the novel algorithm that was used to get the position of the object in the video which was being captured by the camera in the robotic system, in this method the captured video is converted from RGB colour space to HSV for detecting the object based on the HSV value of the object. For this, we need to set a range of HSV values for the object to be identified such that, the object within that specified range of HSV values is detected. A morphological filter[11] is used, which on using the median filter rounds off the large structures to remove the small structures and grow back the remaining structures by the same amount, this process is also called as "shrink and grow process". In the gaussian blur operation [12], the "Gaussian Filter" is a low pass filter that removes the high-frequency component. The 'Hough circles' method [13] from the 'OpenCV' library was used to detect the circular object within the specified range of HSV values in the video and a circle is drawn around the detected circular object. This method draws a red circle around the circular object if it is too close to the robot or closer than a threshold value which we have set up by setting up the threshold radius of the circle drawn around the object and when the radius of the circle drawn is bigger than the threshold radius value, the circle's colour will be red and when the radius is below the threshold radius value, the circle's colour will be green. The robot falls into a state of rest irrespective of the $x_{angle}$ if the object is close or when the circle is red in colour and if the object is away from the robot or when the circle is green in colour, the robot computes the horizontal position or $x_{angle}$ which is the horizontal pixel value at which the mid-point of the object is located in the video or we can simply say that $x_{angle}$ is the horizontal position of the mid-point of the detected object, measured in pixels and this computed value of $x_{angle}$, the colour of the circle around the object detected along with the left sensor and the right sensor values are fed to the ANNs and the outputs of the ANNs which are linear and angular velocities are further passed to the PI controller to control and make the robot move closer to the object to fall into a state of rest. This process continues and keeps the robot moving closer to the object until it reaches close enough to the object where the circle becomes red. This novel method that we have used here was inspired by the colour-based image segmentation method. For a better understanding, Fig. 3, Fig.4 and Fig. 5 shows how the horizontal position of the object is computed from the video that the camera records. Those figures also show how the proposed object detection method detects the object in the video frame. In Fig. 3 the object is located far from the robot at $x_{angle} = 90$, in Fig. 4 the object is located far from the robot at $x_{angle} = 210$ and in Fig. 5 the object is located close to the robot at $x_{angle} = 165$. In Fig. 3 and Fig. 4, the respective circles around the respective objects are green in colour as the objects are located far enough and the respective radiuses of the respective circles around the respective objects in the respective pictures are lesser than the threshold value of radius that was have set to detect if the object is close or far from the robot and in Fig. 4 the circle is red in colour as the object is far and the radius is higher than the threshold value of the radius that was set. The circles around the objects in the figures are drawn using cv2.circle() method from the "OpenCV" image processing library and the $x_{angle}$ is printed beside the detected object using cv2.putText.

OpenCV is an open-source library which consists of a lot of computer vision algorithms. OpenCV has a modular structure, it consists of several shared or static libraries, the modules that are available in OpenCV are 'core functionality', 'image processing', 'video analysis', 'camera calibration and 3D reconstruction', '2D features framework', 'object detection', 'high-level GUI', 'video I/O' and some other helper modules [14] [15], such as Google test wrappers, FLANN and other modules [16].

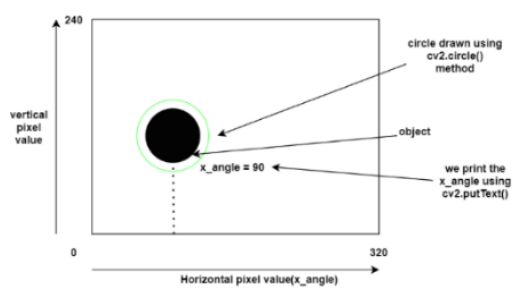
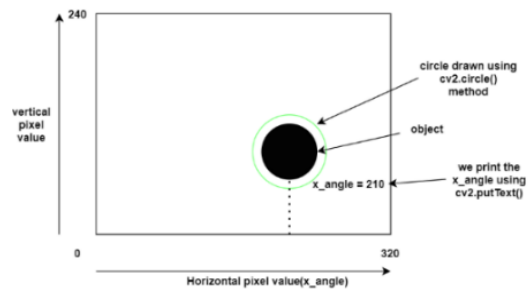

Fig. 3. Object detection in the video frame when the object is not close and x_angle = 90

Fig. 4. Object detection in the video frame when object is not close and x_angle = 210

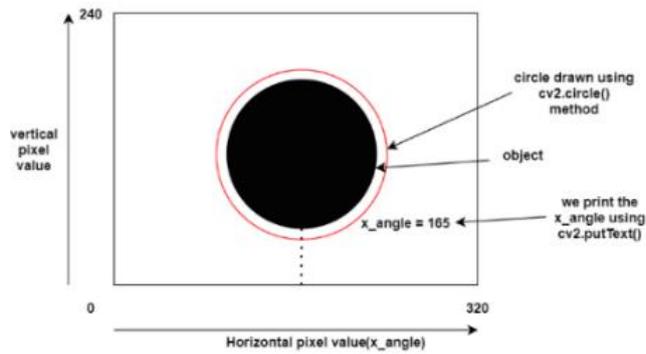

**Fig. 5. Object detection in the video frame when the object is close and x_angle = 165**

We have taken the colour spaces based image segmentation algorithm [17] as a reference algorithm to build the algorithm shown in Fig. 2, to get the position of the ball apart from detecting it, this can also be done using 'Haar cascade classifiers' object detection algorithm [18] in which we detect the object by training the cascade classifier with negative images and positive images but for this particular problem, we have chosen the algorithm shown in Fig. 2. So, using this we get the midpoints of the circle drawn around the object, from which we get the $x_{angle}$ and whether the object is close or far from the robot, these values are passed on to the artificial neural network from which we get the output for the motion of the robot.

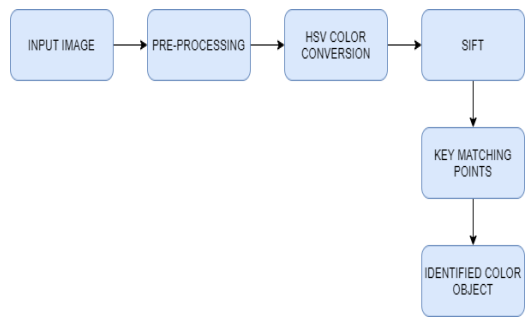
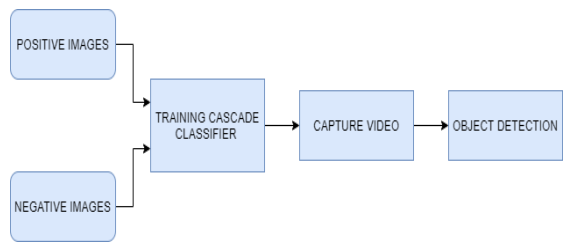

**Fig. 6. Colour based image segmentation**     **Fig. 7. Haar feature based cascade classifier**

Fig. 6 and Fig. 7 show the 2 main algorithms that are used for object detection, the first one is 'colour-based object detection' and the second one is 'Haar feature-based cascade classifier'. The Colour based detection algorithm doesn't need any data for learning whereas the 'Haar feature-based cascade classifier' need negative images which are those images in which the object that needs to be detected isn't present and positive images are those images in which the object that needs to be detected is present.

In 'Haar feature-based cascade classifiers' positive and negative images are fed into the cascade classifier, positive images are those images which consist the object that needs to be detected and the negative image doesn't consist the object that needs to be detected. After training the classifier we can detect the object in a video or an image. Colour spaces-based image segmentation needs no training data, it detects the objects by the colour of the object, we just set the HSV limits of the object and the object is detected based on its colour.

## 3.3. Artificial neural networks for mobility planning

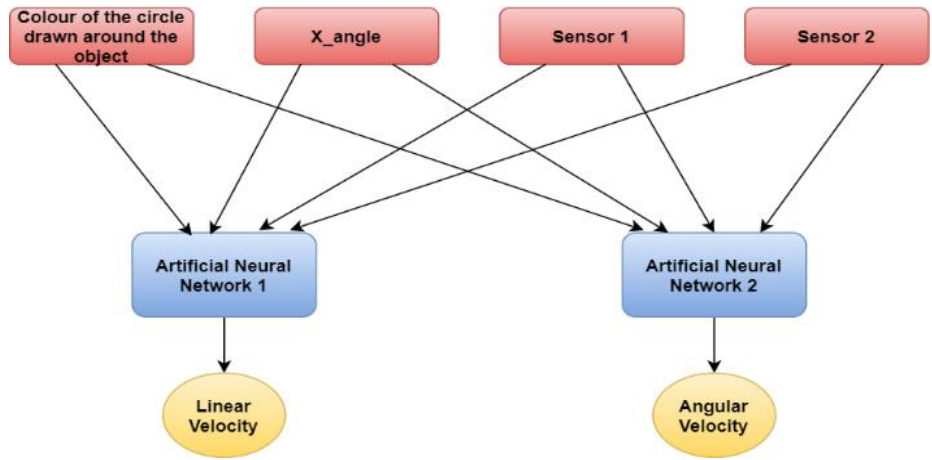

**Fig. 8. Artificial neural networks for computing linear and angular velocity**

Artificial Neural Networks or ANNs or connectionist systems are computing systems which are inspired by biological neural networks that constitute animal brains. These computing systems are trained to give an appropriate output. The weights of the artificial neural network get adjusted by backpropagation algorithm depending upon the loss computed by the loss function, the training data consists around 500 or 1000 or 10,000 rows of data samples depending upon the complexity of the problem [19]. In this problem, we are using 2 artificial neural networks which are deep neural networks or DNNs with 2 hidden layers and the input features are the distance of the obstacle from the left sensor(sensor1), the distance of the obstacle from the right sensor(sensor2), $x_{angle}$ of the object and the colour of the circle around the object which is red or green. The chosen input features are those that are required for the ANNs to predict the linear and angular velocities. The outputs that are computed are linear velocity and angular velocity, instead of predicting linear velocity and angular velocity we could also predict the direction of the motion of the robot, which is left or right or forward and in that case we wouldn't need a PI controller to control the motion of the robot but, the motion is jerky in that case so, we use artificial neural networks to predict linear and angular velocities. Fig. 8 shows how the ANNs were used. If the object is lost, the last updated $x_{angle}$ is taken into consideration and there is a very high probability for the robot to find the lost object after it gets close to the last updated $x_{angle}$.

Using deep neural networks in this type of problems eliminate the need to design the non-linear motion planning equations and this reduces the complexity of the motion (or mobility) planner. Since the robot's design using two artificial neural networks gets simpler, the time taken for motion planner design will get reduced. The artificial neural networks that we have used gives us the linear velocity and the angular velocity as outputs. $v$ and $\omega$ can be converted to $v_r$ and $v_l$ which are the velocity of the right wheel and the velocity of the left wheel, respectively [20]. Convolutional neural networks can also be used in this particular problem but that would increase the computational cost since an expensive graphical processing unit is required to make to the autonomy vision based.

TABLE. 1. The architecture of the deep neural networks used for computing linear velocity

| Layers | Neurons | Activation Function |
|---|---|---|
| Input Layer | 4 | |
| Hidden Layer 1 | 6 | RELU |
| Hidden Layer 2 | 6 | RELU |
| Output Layer | 1 | Linear |

TABLE. 1 describes the architecture of the DNNs or ANNs that we have used, and the activation functions used in every layer. For compiling the DNN, we have used 'Adam' optimizer which is an advancement or update of the 'RMSprop' optimizer[21] and 'MAE' or Mean Absolute Error as loss function [22] to update the weights using backpropagation [23]. The metric that we used while compiling the DNN is 'Loss'. Since we are getting linear velocity in the output, this becomes a regression problem and the 'Accuracy' metric cannot be used in regression problems.

In ANNs when we use Adam optimizer the weights are updated as shown below -

$$w^{t+1} \leftarrow w^t - \eta * \frac{m_t}{\sqrt{v_t + \epsilon}} \qquad (1)$$

Where $w^{t+1}$ and $w^t$ are updated weights and weights before updating, $\eta$ is the learning rate, $m_t$ and $v_t$ are updated biased first-moment estimate and updated bias second raw moment estimate. $\epsilon$ is a small scalar, used to prevent division by 0.

The below equation shows us the loss function that we used in this DNN, which is 'MAE'-

$$Q = \frac{1}{n}\sum_{i=1}^{n}|y_i - y_i'| \qquad (2)$$

Where $y$ is the actual value and $y'$ is the predicted value.

TABLE. 2. The architecture of the deep neural networks used for computing angular velocity

| Layers | Neurons | Activation Function |
|---|---|---|
| Input Layer | 4 | |
| Hidden Layer 1 | 6 | RELU |
| Hidden Layer 2 | 8 | RELU |
| Output Layer | 1 | Linear |

TABLE. 2 describes the architecture of the DNNs or ANNs that we have used for predicting angular velocity and the activation functions used in every layer. For compiling the DNN we have used 'Adam' optimizer mentioned in(1) which is an advancement or update of 'RMSprop' and 'MSE' or Mean Squared Error as loss function [24] to update the weights since it gave us better results. The metric that we used while compiling the

DNN is 'Loss'. Since we are getting angular velocity in the output, this becomes a regression problem and the 'Accuracy' metric cannot be used in regression problems.

We tried to predict the linear velocities and angular velocities using a single neural network architecture but the predictions, after training it with the collected dataset were inaccurate in few time steps and we also tried different architectures with two neural networks to predict the linear and angular velocities, the trial and error method has led us to the given architectures shown in TABLE. 1 and TABLE. 2.

The below equation shows the loss function that has used in this DNN, which is 'MSE'-

$$Q = \frac{1}{n}\sum_{i=1}^{n}(y_i - y_i')^2 \qquad (3)$$

Where $y$ is the actual value and $y'$ is the predicted value.

3.4. Design of the Digital PI Controller

Almost every well-designed mobile robot uses a PID controller [25][26], we can use a P controller, PI controller, PD controller and PID controller depending upon the problem that we have. For this problem we are using a PI controller since we need to control only the velocities[27], in the PI controller there are 2 constants $k_p$ and $k_i$. These constants can be tuned by trying different values until we see the step response graph with a decent rise time and no peak overshoot [28]. The P controller which is also known as a proportional controller is used to amplify the error signal and thus the amplified error signal can be detected and $k_p$ is always greater than 1 since the system needs to be stabilized even for a small error and using P controller, which amplifies the error, the output can be stabilized quickly and the D controller control action increases when the error changes at a faster rate and when the change in error is 0 the control action from the D controller is 0, it provides kickstart to prevent overshooting. The I controller is used to remove the long-term error and the offset in the control system, the integral controller sums the error value over a period of time which is the period of time from 0 seconds to the present time step [29]. The control output given from the PID controller is given as-

$$u(t) = k_p * error + k_d * \frac{d(error)}{dt} + k_i * \int_0^t error(\tau)d\tau \qquad (4)$$

Let us say the control output of the PI controller at a given time, proportional gain, and integral gain as $u(t)$, $k_p$ and $k_i$ respectively.

Let us take the linear velocity error as $v_{error} = v_{desired} - v_{actual}$. The equation for the control input for linear velocity control is given as-

$$u_v(t) = k_p * v_{error} + k_I * \int_0^t v_{error}(\tau)d\tau \qquad (5)$$

Using this controller, we ran simulations and set the appropriate value of $k_p$ and $k_i$, if the value of $k_p$ is very large then the system oscillatory hence we need to choose a value of $k_p$ where the controlled graph just reaches below the overshoot value.

The Laplace transform of the controller equation is given as-

$$u(s) = E(s) * (k_p + \frac{k_i}{s}) \qquad (6)$$

Similarly, we design the PI controller for angular velocity where $\omega_{error} = \omega_{desired} - \omega_{actual}$ is the angular velocity error. The control input for angular velocity control is given as-

$$u_\omega(t) = k_p * \omega_{error} + k_i * \int_0^t \omega_{error}(\tau)d\tau \qquad (7)$$

Let us take the controlled linear and angular velocities as $v_c$ and $\omega_c$ and let us take the actual linear and present angular velocities at time t as $v_{actual}$ and $\omega_{actual}$. Now the controlled linear and angular velocities look like –

$$v_c = v_{actual} + u_v(t) \qquad (8)$$

$$\omega_c = \omega_{actual} + u_\omega(t) \qquad (9)$$

3.5. Converting υ and ω to $v_r$ and $v_l$

$v_c$ and $\omega_c$ which are given by the PI controller need to be converted into $v_r$ and $v_l$. After conversion, the controller will make the right wheel and the left wheel of the robot move, as per the computed values, $v_r$ and $v_l$.

Let's say the length of the robot with 2 moving or driving wheels and one driven wheel or freewheel in the front which has no constraints to its rotation to be 'L' and the radius of both the driving wheels is 'R'.

$$v = \frac{v_l + v_r}{2} \qquad (10)$$

$$\omega = \frac{R}{L} * (v_r - v_l) \qquad (11)$$

After solving equation 10 and equation 11 for $v_l$ and $v_r$ [14] we get -

$$v_l = \frac{2v - \omega L}{2R} \qquad (12)$$

$$v_r = \frac{2v + \omega L}{2R} \qquad (13)$$

Hence, $v_r$ and $v_l$ can be written as-

$$\begin{bmatrix} v_r \\ v_l \end{bmatrix} = \frac{1}{R} * \begin{bmatrix} 1 & \frac{L}{2} \\ 1 & -\frac{L}{2} \end{bmatrix} * \begin{bmatrix} v \\ \omega \end{bmatrix} \qquad (14)$$

## 4. HARDWARE DESIGN OF THE ROBOT

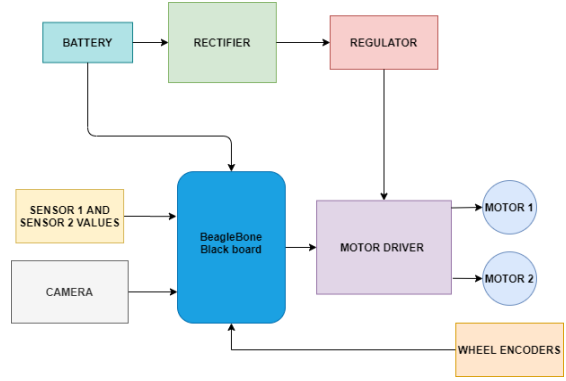

**Fig. 9. Hardware of the robot**

Fig. 9 shows the hardware of the robot, we have used a Beagle Bone Blackboard [31] connected to 2 sensors and one camera at the input and at the output, there are 2 motors interfaced with the Beagle Bone Blackboard. The sensors and the camera give their inputs to the Beagle Bone Blackboard which is a minicomputer to control the robot. The board takes the inputs, computes the output according to our algorithm and the outputs, in this case, are $v\ and\ \omega$. The wheel encoders are used to count the number of revolutions of the wheel of the robot using which we can calculate the distance travelled by the robot and the path of the robot.

Let us say $D_l$, $D_r$, $D_c$ are the distances travelled by the left wheel, distance travelled by the right wheel and distance travelled by the centre of the robot. We get $tick_l, tick_r$ from the wheel encoders and 'N' is the number of ticks per revolution.

The equations that tell us the distance travelled by the robot are given as -

$$number\ of\ revolutions = \frac{\Delta tick_l}{N} \qquad (15)$$

$$D_l = \frac{2\pi R * \Delta tick_l}{N} \qquad (16)$$

$$D_r = \frac{2\pi R * \Delta tick_r}{N} \qquad (17)$$

$$D_c = \frac{D_r + D_l}{2} \qquad (18)$$

The position of the robot can be calculated using $D_l, D_r, D_c$ with the help of the below equations –

$$x_{new} = x_{old} + D_c * \cos a \qquad (19)$$

$$y_{new} = y_{old} * D_c * \sin a \qquad (20)$$

$$a_{new} = a_{old} + \frac{D_r - D_l}{L} \qquad (21)$$

In the above equations $x_{new}$, $y_{new}$ and $a_{new}$ are the new X position, the new Y position and the new angle made by the robot with the x-axis, respectively. From the above equations $x_{old}$, $y_{old}$ and $a_{old}$ are the old x position, the old y position and the old angle made by the robot with the x-axis, respectively.

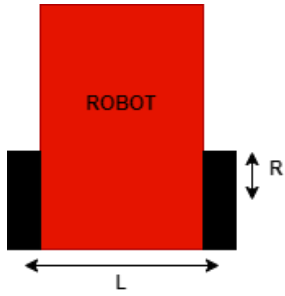
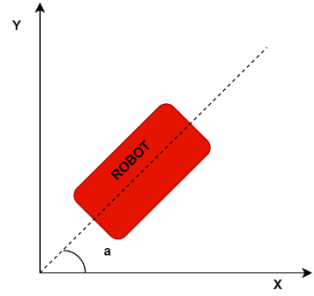

**Fig. 10. Mechanical structure of the robot**

**Fig. 11. Alignment of the robot in X, Y co-ordinates**

The mechanical design of the robot is simple since it is just a 2 driven wheeled robot. The 2 main design specifications that influence the speed of the wheels are the radius or 'R' and length or 'L' of the robot so, we need to be very selective in choosing the length and the radius. The mechanical structure of the robot is given below in Fig. 10 and Fig. 11 shows us the orientation of the robot in 2D, 'a' is the angle between the x-axis and the imaginary axis of the robot shown in Fig. 10. The robot moves in the 2D XY plane.

## 5. EXPERIMENTAL RESULTS

### 5.1. Experimental Setup

For conducting this experiment, we have chosen the popular smiley faced yellow colour ball which resembles a yellow coloured circular object and we have used the "TensorFlow" library [32] in python to build the deep neural networks which are required to predict the linear and angular velocity. The data with which we need to train the ANNs or DNNs were scaled for good train and test performance of the artificial neural networks, we used 'mean absolute error' and 'mean squared error' as the loss functions to compute the loss and adjust the weights through backpropagation. As mentioned earlier the sensors which we used are ultrasonic sensors and it measures the distance of the obstacle in centimetres.

### 5.2. Data Generation

The data to train and test the deep neural network for the given problem can be generated by recording the $x_{angle}$, the value from the left sensor, the value from the right sensor, the colour of the circle around the detected object, linear and angular velocities at every time step by manually controlling the robot, during which we should control the robot in such a way that it follows the object and avoids the obstacles, which enables us to record the data required since, this data is being generated by manually controlling the robot, the behaviour of the robot, after deploying, completely depends upon how well we have manually controlled the robot during the data generation period. The robot behaves akin to how it behaved during data generation when the robot was manually controlled to generate the data required to train the ANNs. This method of data generation was proposed in [30].

### 5.3. Object Detection

We have inspected our robotic system, in the first part of the inspection we are going to discuss those results that we got from the novel algorithm for tracking the position of the object, which was inspired from the standard 'colour space-based detection'. We have used a yellow ball as an object and we set the range for the yellow colour to be detected by mentioning the H, S, V limits.

We set the video captured to 240 pixels vertically and 320 pixels horizontally so, $x_{angle}$ varies from 0 to 320 depending upon the position of the object in the video that's being captured and whenever the object is close to the robot the circle is drawn in red. For the $x_{angle}$ values going below 80 or 100 the object is said to be moving to the left side of the robot, for values in between 80 or 100 and 240 or 220 the object is said to be in front of the robot and for values greater than 220 or 240 the object is said to be moving to the right side. The motion of the robot highly depends upon the generated training data. In Fig. 12, a1 and a2 show how the ball is detected when it is at the left side, b1 and b2 show how the ball is detected when it is placed at the right side, c1 and c2 show how the ball is detected when it is placed in the front and figure d1 and d2 show the change in colour of the circle around the object, from green to red when the ball is very close to the robot.

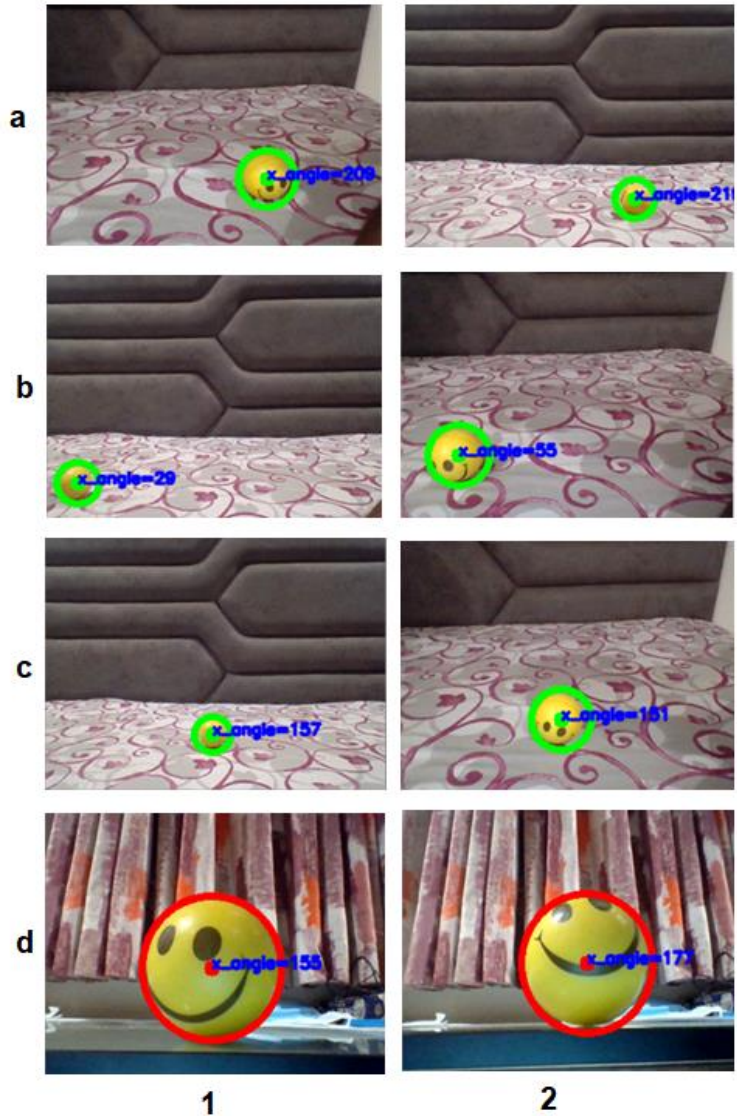

Fig. 12. Results of object detection

The value of $x_{angle}$ which we could see on the pictures is given as an input to the artificial neural networks along with the colour of the circle and values from the sensors which tell us the position of the obstacles and then the output is computed. We need to train the artificial neural networks with scaled data before using it.

## 5.4. Performance of the ANNs

We trained the ANN with a dataset of 5000 rows and the results which we got were quite accurate. We used two multilayer neural networks for this problem as described in 3.3, the ANNs give us two output which is $v\ and\ \omega$. We compared the actual $v\ and\ \omega$ with the predicted $v'and\ \omega'$ to show the performance of the model while working in real-time and we plotted the training loss and validation loss which confirms the reliability of the model.

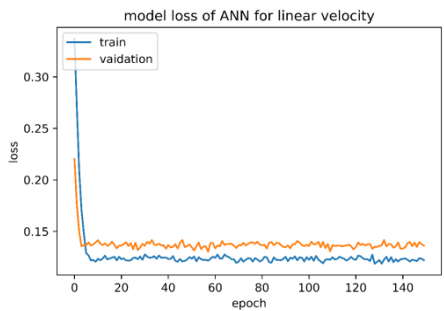 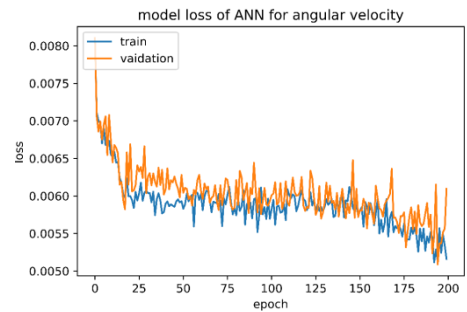

**Fig. 13. Loss of the neural network 1(neural network which predicts linear velocity) in every epoch**

**Fig. 14. Loss of the Neural Network 2(neural network which predicts angular velocity) in every epoch for predicting angular velocity**

Fig. 13 and 14 show us the performance of the artificial neural networks with outputs of linear and angular velocity while training and the validation of the model for $v$ and $\omega$. The loss of the model converges to a value lower than 0.10 and 0.0055 for linear and angular velocity predicting ANNs, respectively, which tells us the reliability of the model. Since this is a robotic system we need an excellent performance from the motion planner which is a model with two artificial neural networks in this case. The desired values of linear velocity and angular velocity are computed by the artificial neural networks which are given to the PI controller to get the controlled linear velocity and angular velocity. The ANNs that we have used give us accurate enough results to avoid obstacles. The ANNs that we have used are not 100% accurate because it is based only on the current sense data but, the accuracy of the model and the loss of the model are enough to avoid obstacles extraordinarily. Before training the artificial neural network, we have scaled the training and testing data using "StandardScalar" method from the sci-kit learn library [32].

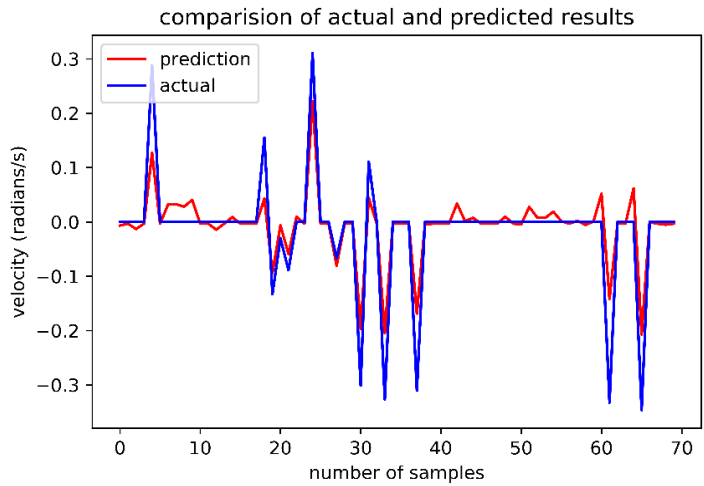

**Fig. 15. Performance of the ANN model in in predicting angular velocity during sudden appearance of the obstacles**

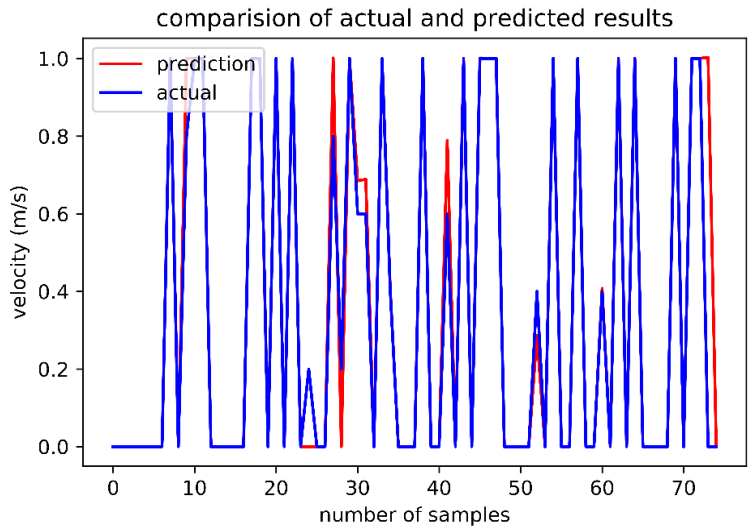

**Fig. 16. Performance of the ANN model in predicting linear velocity during sudden appearance of the obstacles**

Fig. 15 and Fig. 16 show us the performance of the ANNs in predicting angular and linear velocity, respectively, when obstacles appear suddenly. We have tested both the multilayer neural networks on a dataset in which the left sensor value and the right sensor value which are the distances from the obstacles towards left and right, respectively, change suddenly to a very low value which only happens when the obstacles come closer to the robot all of a sudden(dynamic obstacles). We did this to test the reliability of the model during the sudden appearance of the obstacles in the surroundings. The maximum linear speed of the robot is 1 m/s, we modified the generated data to get the training data by making all the values under the linear velocity column $\geq 1 m/s$ as $1\ m/s$ and trained the model with this modified data to predict linear velocity with maximum linear velocity as $1\ m/s$, this was done for getting a clearer graph. The performance of the ANNs shows us the reliability of the model in sudden changes in the surroundings. Fig. 15 and Fig. 16 show the predicted plot and the actual plot in the sudden appearance of the obstacles. These results show us that the ANNs respond to the sudden appearance of the obstacles.

5.5. Evaluation of the PI controller

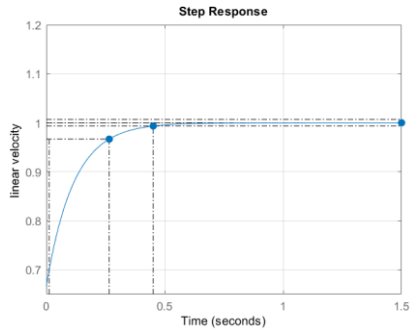

**Fig. 17. Step response of the PI controller for linear velocity**

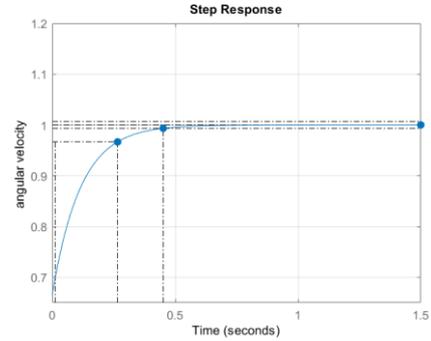

**Fig. 18. Step response of the PI controller for angular velocity**

Fig. 17 and Fig. 18 show us the step responses of the PI controller for linear and angular velocity, respectively. Since we used the same control design for both angular and linear velocities the response looks similar. We have tuned the controller adjusting the $k_p$ and $k_i$ constants and we got the settling time at 0.451 seconds and raise time at 0.254 and the controller needs to attain stability very quickly because failing to do so the robot may slam into the obstacle. The $k_p$ and $k_i$ gains were set by using the trial and error method. This quick response from the robot keeps it away from the obstacles and make it follow the person continuously. The rise time and settling time can further be reduced but doing so it will increase jerks and the robot may collapse hence the rise time and settling time needs to be suitable for the structure of the robot.

5.6. Testing the Robot in 3 different environments

We tested the robot by setting up 3 environments for the robot, in which we have used a yellow coloured ball which was hung by a thread, as the moving object that needs to be followed. The thread by which the ball was hanging, was held by a person, to move it in the environment. We have used rectangle-shaped obstacles to test the robot's obstacle avoidance ability. We have trained the robot with low speeds for better graphs, the speed can be increased by training the linear velocity predicting ANN with higher linear speeds. The obstacles were rectangle-shaped with varying dimensions. In the first environment, we have used 3 obstacles at different positions in the environment to test the robot. The robot avoided the obstacles while following the object accurately in all 3 environments. During the movement of the robot in the different test environments, using the wheel encoders, we recorded the positions (x and y coordinates) of the head of the robot in the XY-plane, to plot the path of the robot.

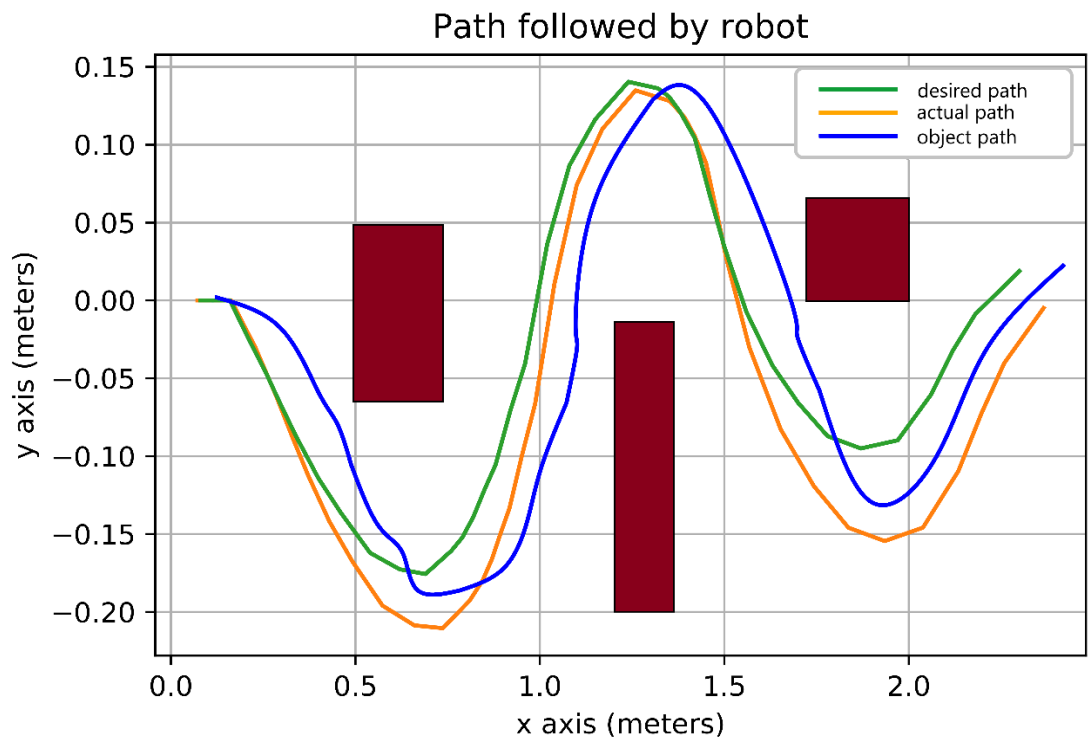

**Fig. 19. Path followed by the robot in environment 1**

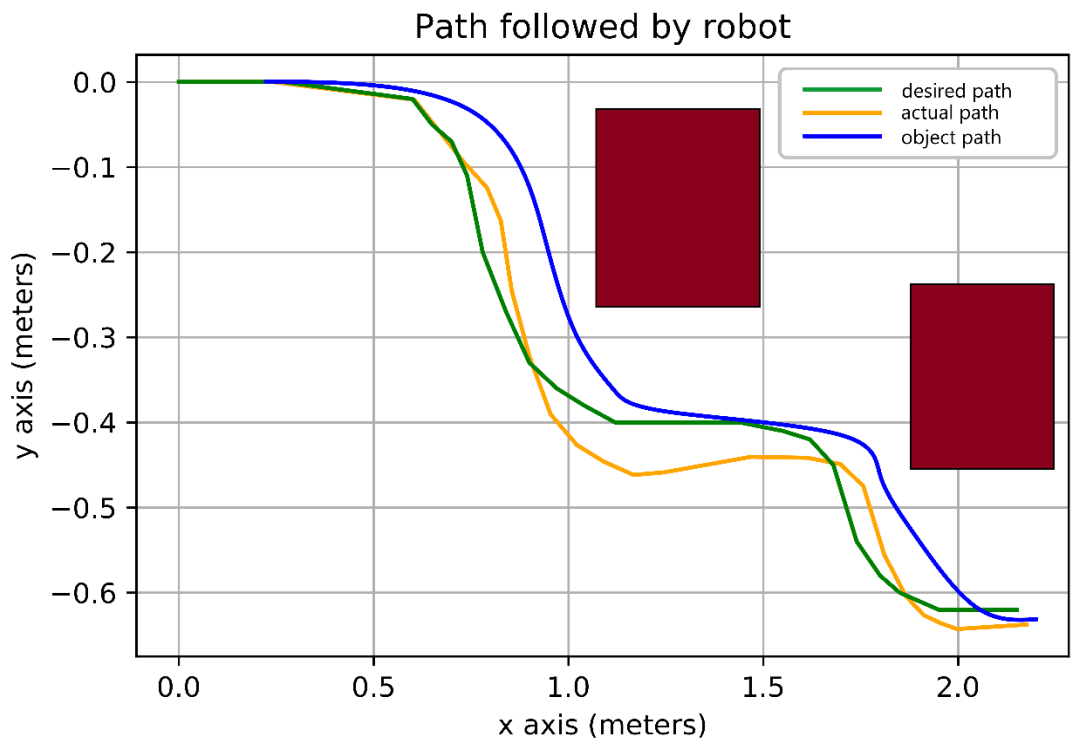

**Fig. 20. Path followed by the robot in environment 2**

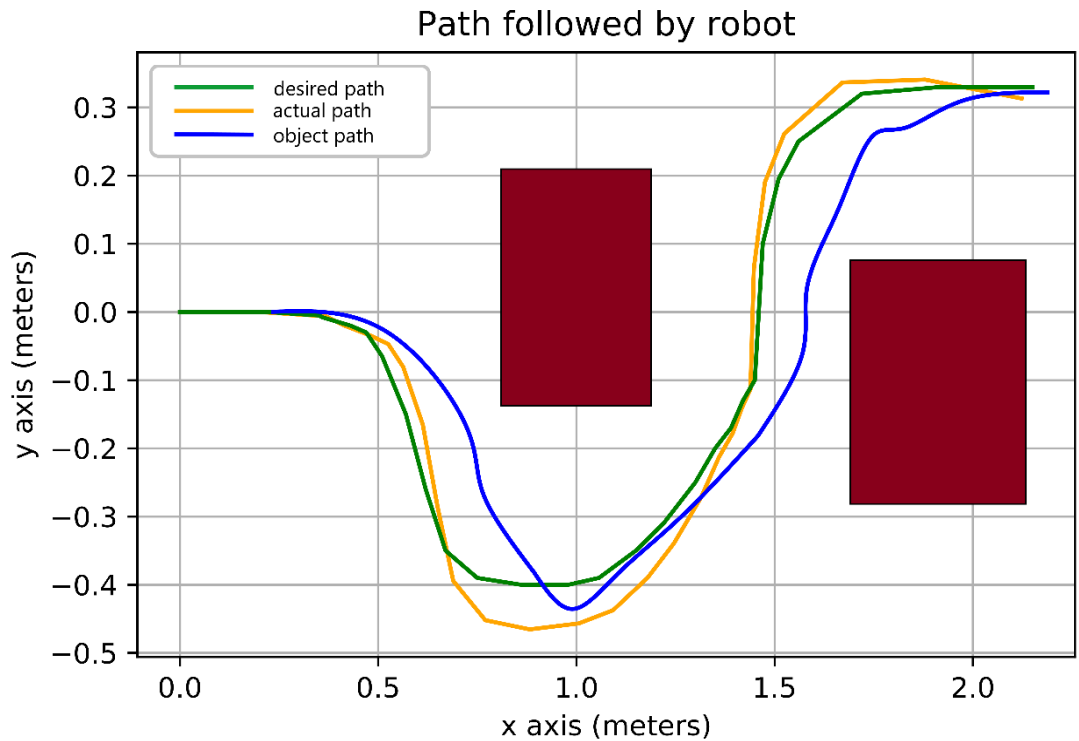

Fig. 21. Path followed by the robot in environment 3

Fig. 19 shows the actual path of the robot, the desired path of the robot, the path of the object, and the obstacles in environment 1. The actual path is the path followed by the robot when the robot's linear and angular velocities are the controlled linear and angular velocities given out by PI controller in real-time and desired path is the path that is computationally generated by multiplying the time period of every time step, which is 0.5 seconds with the linear and angular velocities given out by the respective neural networks in that particular time step. The robot took 19 seconds to move in the actual path in environment 1, which was just over 2.3 meters, by following the object and by avoiding the obstacles. Similarly, Fig. 20 and Fig. 21 show the robot's path and the object's path in environment 2 and environment 3, respectively. The robot, while following the object, can also avoid dynamic obstacles. The comparison of actual and predicted linear velocities in an environment with dynamic obstacles are shown in Fig. 14 and the comparison of actual and predicted angular velocities in an environment with dynamic obstacles are shown in Fig. 15. The environment 2 was 2.0 to 2.5 meters long and the robot, on following the object, covered it in 12.5 seconds. The environment 3 was also 2.0 to 2.5 meters long and the robot, on following the object, covered it in 13 seconds. The object was moved slowly in the environment, which makes the robot move slowly, for obtaining clear graphs. If you have observed the figures, Fig. 19, Fig. 20, and Fig. 21, you can see the robot moving away from the object and the obstacles whenever, the object moves close to the obstacles, which shows the promising performance of the robot.

$v_c$ and $\omega_c$, which are the controlled linear velocity and the controlled angular velocity, are given out from the PI controller in every time step, Fig. 22 and Fig. 23 show us the comparison of the desired linear velocity with the controlled linear velocity and the comparison of the desired angular velocity with the controlled angular velocity of the robot in environment 1, respectively. The controlled linear velocity and the controlled angular velocity is given by the PI controller, which controls the movements of the robot. The linear velocity was measured in meters per second and the angular velocity was measured in radians per second and each time step is equivalent to 0.5 seconds as the neural networks predict new linear and angular velocity every half second. Similarly, Fig. 24 and Fig. 25 show the comparison in environment 2, Fig. 26 and Fig. 27 show the comparison in environment 3.

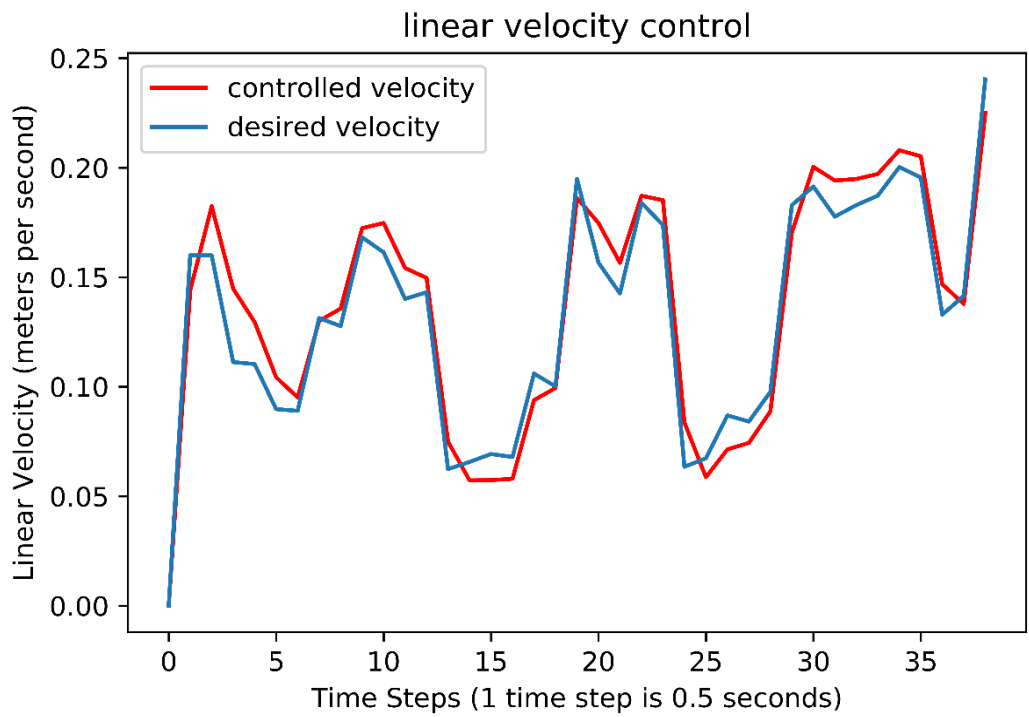

Fig. 22. Linear velocity control of robot in environment 1

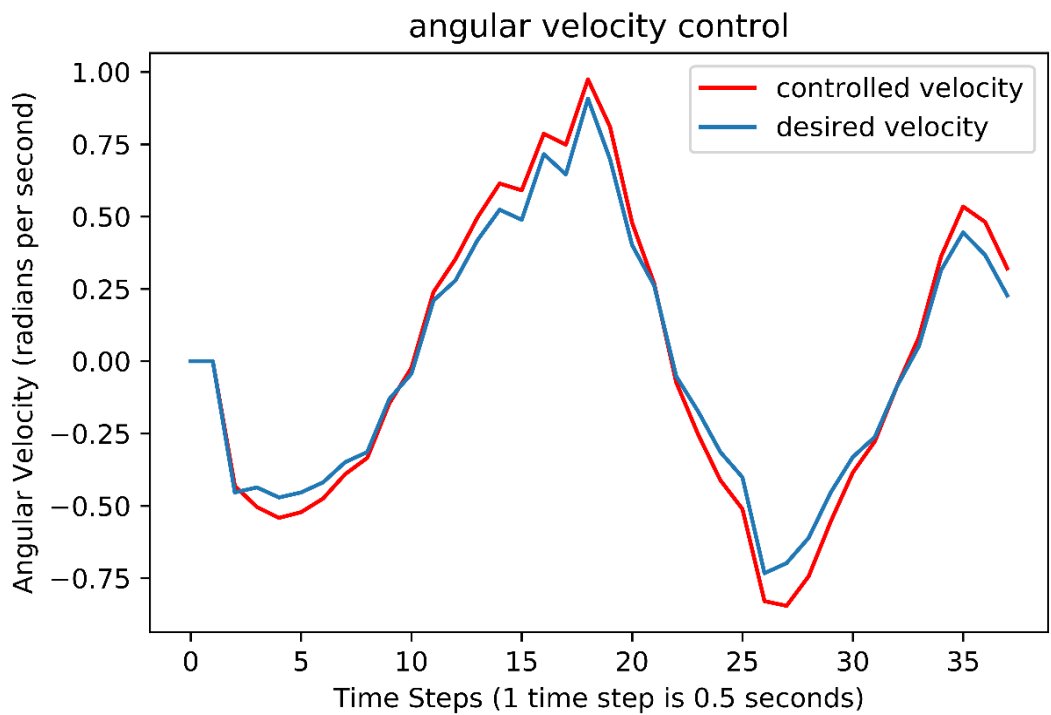

Fig. 23. Angular velocity control of the robot in environment 1

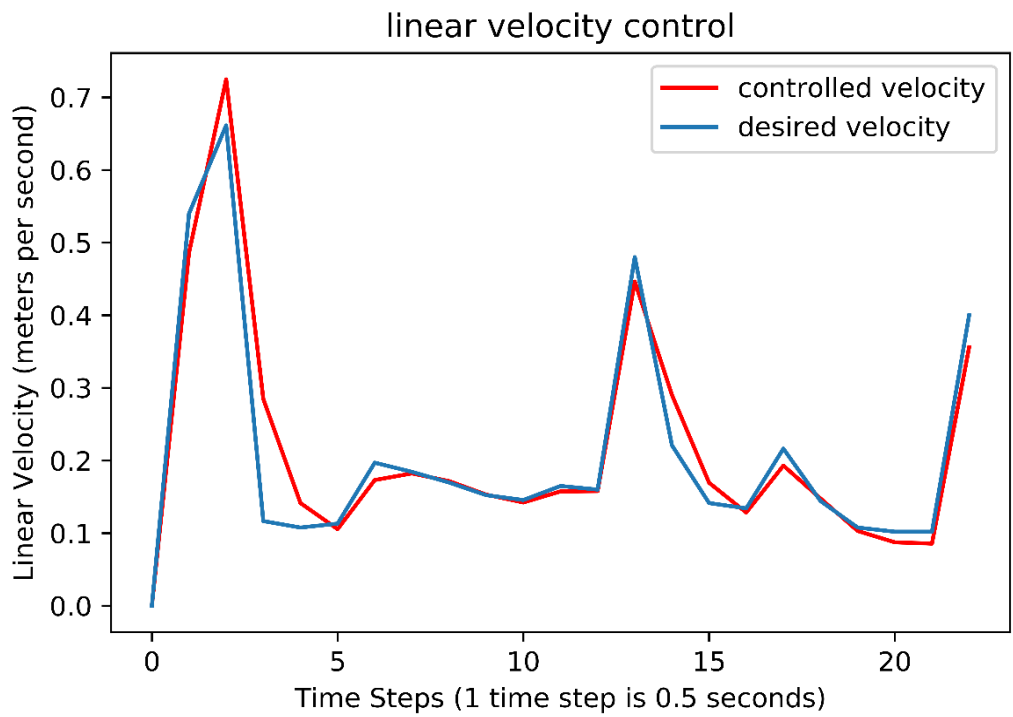

**Fig. 24. Linear velocity control of the robot in environment 2**

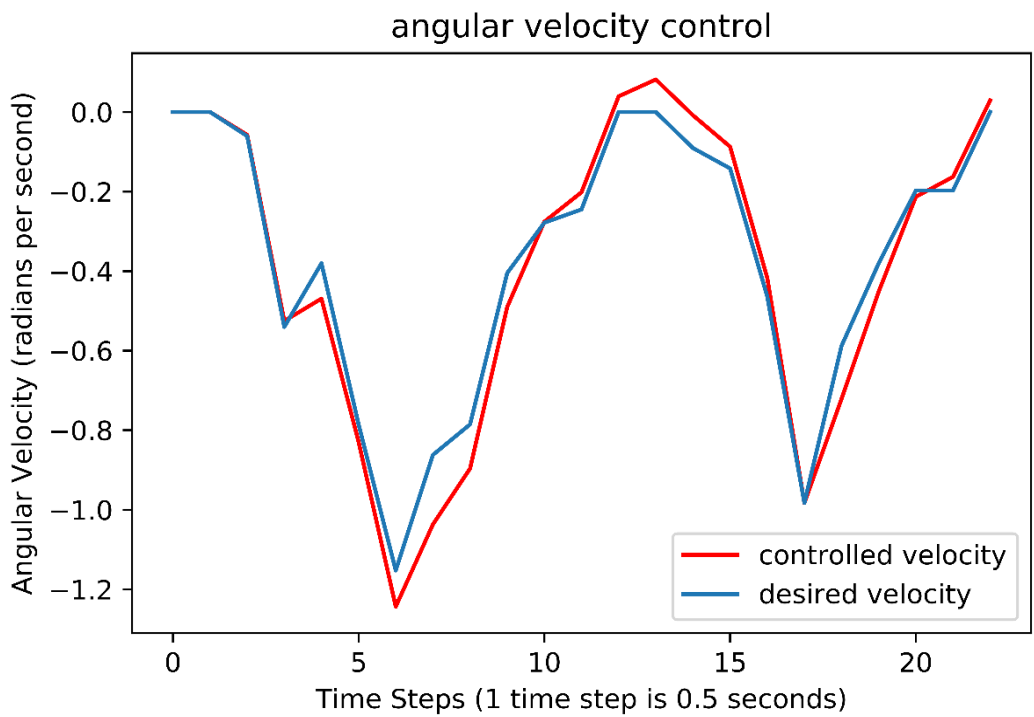

**Fig. 25. Angular velocity control of the robot in environment 2**

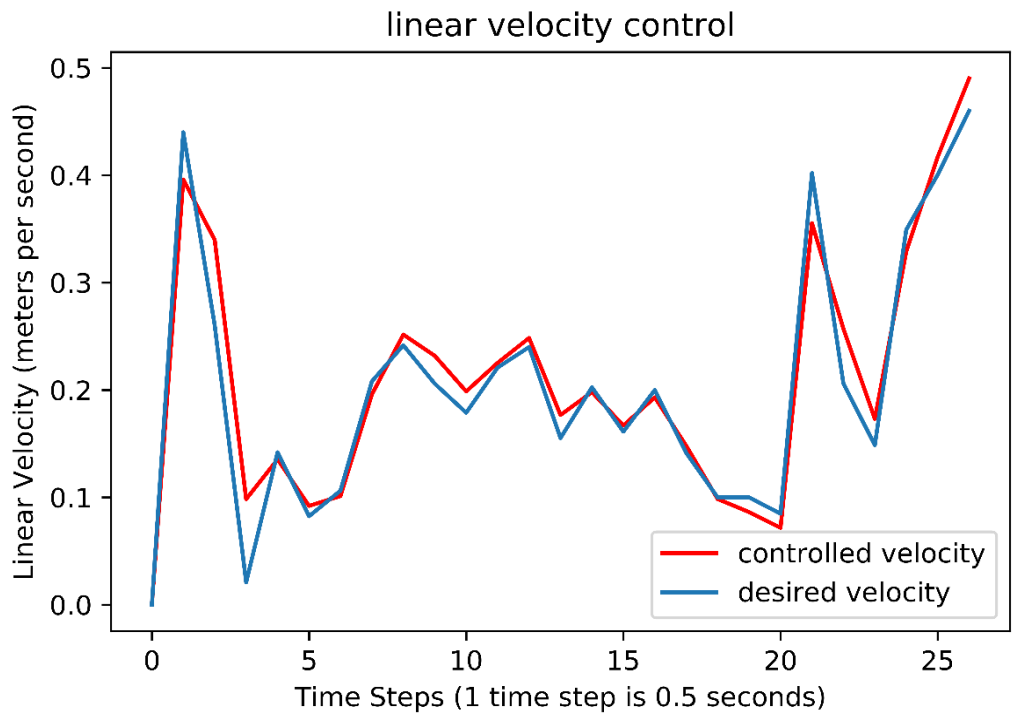

**Fig. 26.** Linear velocity control of the robot in environment 3

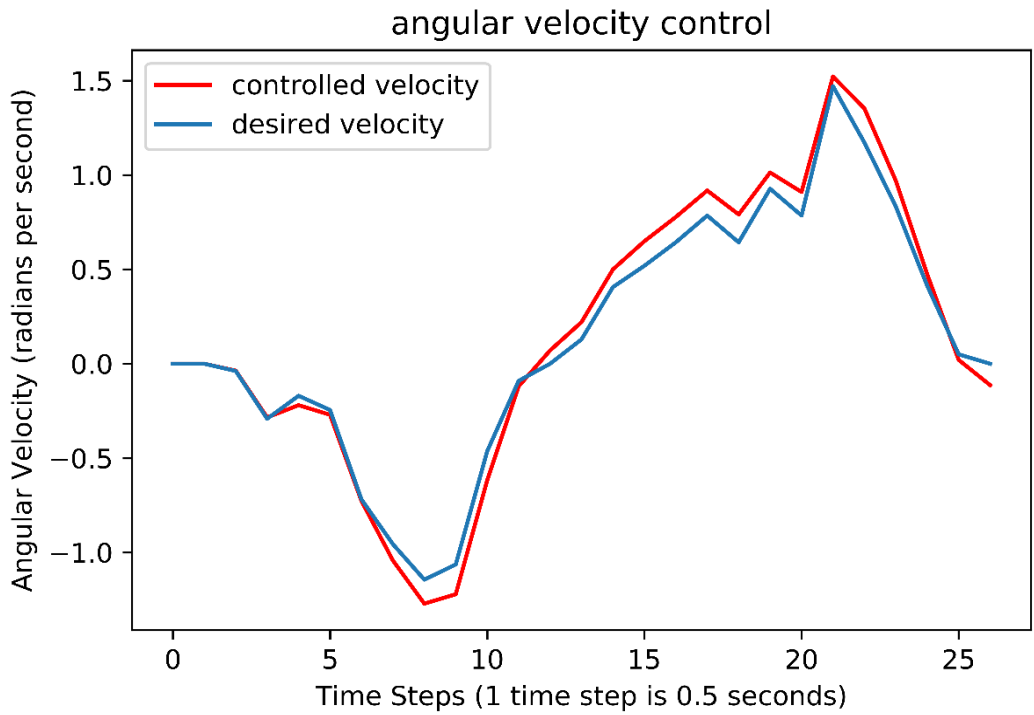

**Fig. 27.** Angular velocity control of the robot in environment 3

## 6. CONCLUSION

In this paper, we propose a robotic system for following objects with avoiding the obstacles using computer vision and artificial neural networks since the technique suggested uses artificial neural networks the complexity of the problem is reduced and motion planning can we achieved as desired by training the neural networks. The main contribution of the paper is the usage of the algorithm in Fig. 2, which was inspired by the standard colour-based object detection algorithm to get the position of the image and the usage of artificial neural networks to plan the motion or mobility of the robot. The proposed methodology eliminates the problems that arise during obstacle avoidance, this method also deals properly with transparent obstacles since we are using ultrasonic sensors to detect the obstacles and it also avoid the sudden appearing obstacles. The PI controller controls the linear and angular velocities accurately in changing the robot's velocities accordingly. The results and the viability of our proposed methodology have been demonstrated successfully and the outcome, which is the position of the robot in every time step shows how accurate the model has worked. The future plan is to train the neural networks in challenging environments and make the robot more reliable and test the robot's performance using LSTMs instead of the standard neural networks, which might give better results.

**Funding**

This research was funded by Chaitanya Bharathi Institute of Technology, Telangana, India.